\documentclass[12pt]{article}

\usepackage{amssymb,amsmath,amsfonts,latexsym,graphicx,hyperref}
\usepackage[numbers]{natbib}

\begin{document}

\begin{center}
\Large \bf  Transformer-Based Wireless Capsule Endoscopy Bleeding Tissue Detection and Classification 

\normalsize \bf An Auto-WCEBleedGen Version 1 Challenge Report By KU Researchers \rm

\vspace{1cm}

%  List of authors.

\large Basit Alawode$\,^a$, \large Shibani Hamza$\,^a$, \large Adarsh Ghimire$\,^{a,b}$, \large Divya Velayudhan$\,^a$

\vspace{0.5cm}

\normalsize

%  List of the affiliations depending on the letters a, b,... used in the list of authors.

$^a$ Khalifa University of Science and Technology, Abu Dhabi, UAE.

$^b$ Halcon, Abu Dhabi, UAE.

\vspace{5mm}

% Email address of the authors. 

Email: {\tt basit.alawode@gmail.com, shibani.h@gmail.com, adarshghimire5@gmail.com, divyavelayudhan@gmail.com}

\vspace{1cm}

\end{center}

\abstract{Informed by the success of the transformer model in various computer vision tasks, we design an end-to-end trainable model for the automatic detection and classification of bleeding and non-bleeding frames extracted from Wireless Capsule Endoscopy (WCE) videos. Based on the DETR model, our model uses the Resnet50 for feature extraction, the transformer encoder-decoder for bleeding and non-bleeding region detection, and a feedforward neural network for classification. Trained in an end-to-end approach on the Auto-WCEBleedGen Version 1 challenge training set, our model performs both detection and classification tasks as a single unit. Our model achieves an accuracy, recall, and F1-score classification percentage score of 98.28, 96.79, and 98.37 respectively on the Auto-WCEBleedGen version 1 validation set. Further, we record an average precision (AP @ 0.5), mean-average precision (mAP) of 0.7447 and 0.7328 detection results. \textbf{This earned us a 3rd place position in the challenge}. Our code is publicly available via \href{https://github.com/BasitAlawode/WCEBleedGen}{https://github.com/BasitAlawode/WCEBleedGen.}

\section{Introduction}
Wireless Capsule Endoscopy (WCE) is a cutting-edge, non-invasive technique for diagnosing small bowel diseases. Its ability to provide comprehensive visualization of the gastrointestinal tract has revolutionized clinical diagnostics. However, the manual frame-by-frame analysis of the vast amount of data collected during WCE procedures poses significant challenges in terms of time and effort, creating a bottleneck for timely and accurate diagnosis.

Artificial Intelligence (AI), particularly its subfield of Deep Learning (DL), offers promising solutions to overcome these limitations. By enabling descriptive, predictive, and prescriptive analyses, AI facilitates the extraction of valuable insights from complex data, which are otherwise unattainable through manual methods. The integration of AI into WCE analysis has the potential to enhance diagnostic accuracy while significantly reducing the workload of medical professionals.

In this work, we present a novel approach based on the Detection Transformer (DETR) architecture to detect and classify frames in WCE data into bleeding and non-bleeding categories. Our model utilizes ResNet50 for feature extraction, a transformer encoder-decoder for detecting bleeding and non-bleeding regions, and a feedforward neural network for classification. Trained end-to-end on the Auto-WCEBleedGen Version 1 challenge training set, the proposed model integrates detection and classification into a unified framework, performing both tasks simultaneously with high efficiency. This approach marks a significant step forward in automating WCE data analysis, minimizing manual effort, and improving diagnostic precision.
\label{sec_intro}

\section{Method}
\label{sec_method}

%%%%%%%%%%%%%%%%%%%%%%%%%%%%%%%%%%%%%%%%%%%%%%%%%%%%%%%%%%%%%%%%%
% Brief write up/summary of the developed pipeline.
%%%%%%%%%%%%%%%%%%%%%%%%%%%%%%%%%%%%%%%%%%%%%%%%%%%%%%%%%%%%%%%%%

Informed by the success of the transformer model in various computer vision tasks, we design an end-to-end trainable model for the automatic detection and classification of bleeding and non-bleeding frames extracted from Wireless Capsule Endoscopy (WCE) videos, as shown in Figure \ref{fig:method}. 

Our model follows the DETR architecture \cite{carion2020end} with a few structural adjustments to the task at hand. Given a tissue frame or image, we utilized ResNet50 \cite{he2016deep} to learn the 2D representation of the tissue (feature extraction). The model then flattens it and adds positional encoding before passing it into a transformer encoder. A transformer decoder then takes as input a small fixed number of learned positional embeddings, called object queries, and additionally attends to the encoder output. The output embedding of the decoder is then passed to a shared feedforward network (FFN) that predicts either a detection (bleed/non-bleed and bounding box) or a “background” class. 

\begin{figure}
    \centering
    \includegraphics[width=\linewidth]{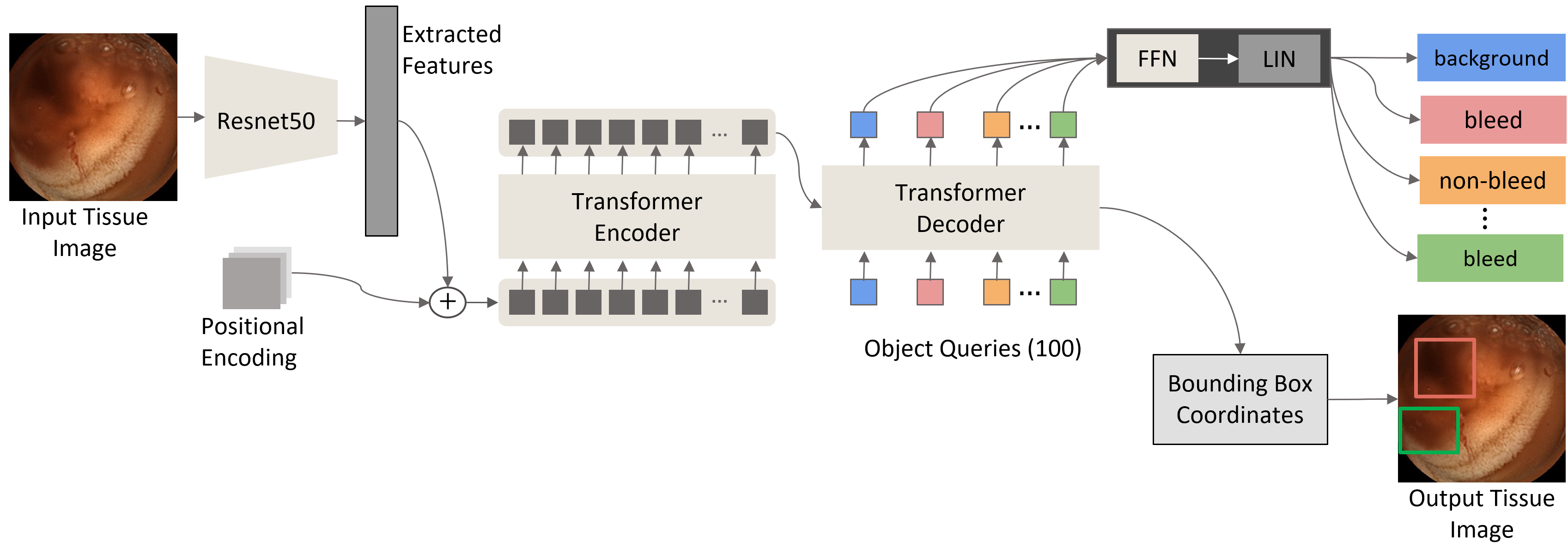}
    \caption{Pipeline of our end-to-end trainable single-model for WCE bleeding tissue classification and detection.}
    \label{fig:method}
\end{figure}

The DETR architecture outputs 80 classes. However, to adapt this for our purpose of classifying bleeding and non-bleeding tissues, we append to, to the output of DETR, a single layer FFN with 3 neurons to predict the probability of the tissue image belonging to the 3 classes of bleed, non-bleed, or background. 

It should be noted that more than one region of a frame can be classified. Additionally, our model outputs the bounding boxes of the regions detected as bleeding or non-bleeding alongside the classification probability. A frame is finally classified as bleeding if at least one region in such frame has been classified as bleeding with a probability greater than 0.5. During inference, the bounding boxes of all the regions classified as bleeding are then superimposed on the frame for visualization purposes.

\subsection{The Training and Test Data}

The training dataset consists of 2618 frames split equally and classified into bleeding (1309 frames) and non-bleeding (1309 frames) WCE frames collected from multiple internet resources, datasets with a wide variety and types of gastrointestinal (GI) bleeding throughout the GI tract, along with medically validated binary masks and bounding boxes in three formats (txt, XML, and YOLO txt). To facilitate our training, we further split the overall training set into train and validation subsets in the ratio 80:20. Both classes are equally represented in the two subsets, resulting in a balanced split. 

The test dataset is an independently collected WCE data containing bleeding and non-bleeding frames of more than 30 patients suffering from acute, chronic, and occult GI bleeding referred at the Department of Gastroenterology and HNU, All India Institute of Medical Sciences, New Delhi, India. 

Both datasets remain the property of the Auto-WCEBleedGen Version V1 Challenge.

\subsection{Training Details}

\subsubsection{Data Preprocessing}
To achieve end-to-end training, we extracted the bounding boxes from the provided dataset XML files and one-hot encoded the classes. These form the output data of our model with the images as the input. To improve the model's robustness, augmentations such as motion blurring, graying, random brightness contrasting, color jittering, random Gammaing, and normalization were applied.

\subsubsection{Loss Function}

Following the DETR training flow, the bipartite matching loss \cite{carion2020end}, which is a weighted combination of the Hungarian matching algorithm \cite{kuhn1955hungarian} loss, the cross-entropy loss \cite{goodfellow2016deep}, and the IOU loss \cite{rezatofighi2019generalized}, is adopted. The Hungarian matching loss ensures a one-to-one matching between predicted outputs, with the bounding boxes and ground truth annotations enabling end-to-end training without the need for hand-designed components like anchor boxes or non-maximum suppression (NMS). The cross-entropy loss is the classification loss measuring the difference between the predicted probability distribution and the true label distribution, penalizing incorrect predictions more severely as they become more confident. The IOU loss captures the overlap between predicted and ground-truth bounding boxes. For our model, the weights of the Hungarian matching loss, the cross-entropy loss, and the IOU loss are 1, 5, and 2, respectively.

\subsubsection{Model Training}
Transformer-based models are known to perform well in the presence of huge amounts of data. As we have only 2618 images for training and validation, training from scratch is impossible. As such, we utilized the trained DETR model with ResNet50 weights available at \cite{detr2020} to initialize all the weights of our model except the last FFN layer that was randomly initialized. Our model is then fine-tuned using the training set.

The model was fine-tuned for 500 epochs using the AdamW optimizer \cite{loshchilov2018decoupled} with a learning rate of $1 \times 10^{-6}$, $1 \times 10^{-5}$, and $5 \times 10^{-5}$ for the ResNet50 backbone, the transformer encoder/decoder, and the final FCN prediction layer, respectively. A weight decay of $1 \times 10^{-4}$ was also adopted. 

\section{Results}
\textbf{Evaluation Metrics:} We have utilized the Accuracy, Recall, and F1-Score metrics to report the classification performance of our model and the average precision (AP@50), mean average precision (mAP), and Recall (@0.5:0.95) for detection performance.

\textbf{Validation Set Results:} Our model's classification and detection performance on the validation set is as shown in Table \ref{tab:perf_compare}. Our model achieves an accuracy, recall, and F1-score classification percentage score of 98.28, 96.79, and 98.37 on the Auto-WCEBleedGen version 1 validation set. Further, we record an average precision (AP @ 0.5), mean-average precision (mAP) of 0.7447 and 0.7328 detection results. Further, we provide sample detection images from the validation set in Figure \ref{fig:val_set} and the corresponding Grad-CAM visualization images in Figure \ref{fig:val_set_cam}.

\textbf{Test Set 1 and Test Set 2 Visual Results:} In Figures \ref{fig:test_set1} and \ref{fig:test_set2}, we provide sample detection images and corresponding Grad-CAM visualization from the test set 1 and 2 respectively. 

\begin{table}
    \centering
    \caption{Model classification and detection performance on the validation set.}
    \begin{tabular}{ c| c c c| c c c}
    \hline
    & \multicolumn{3}{c|}{Classification (\%)} & \multicolumn{3}{c}{Detection} \\
    & Accuracy & Recall & F1-Score & AP@50 & mAP & Recall (@0.5:0.95) \\
    \hline
    Our Model & 98.28 & 96.79 & 98.37 & 0.7447 & 0.7328 & 0.7706 \\
    \hline
    \end{tabular}
    \label{tab:perf_compare}
\end{table}

\begin{figure}[htbp]
    \centering
    \includegraphics[width=\linewidth]{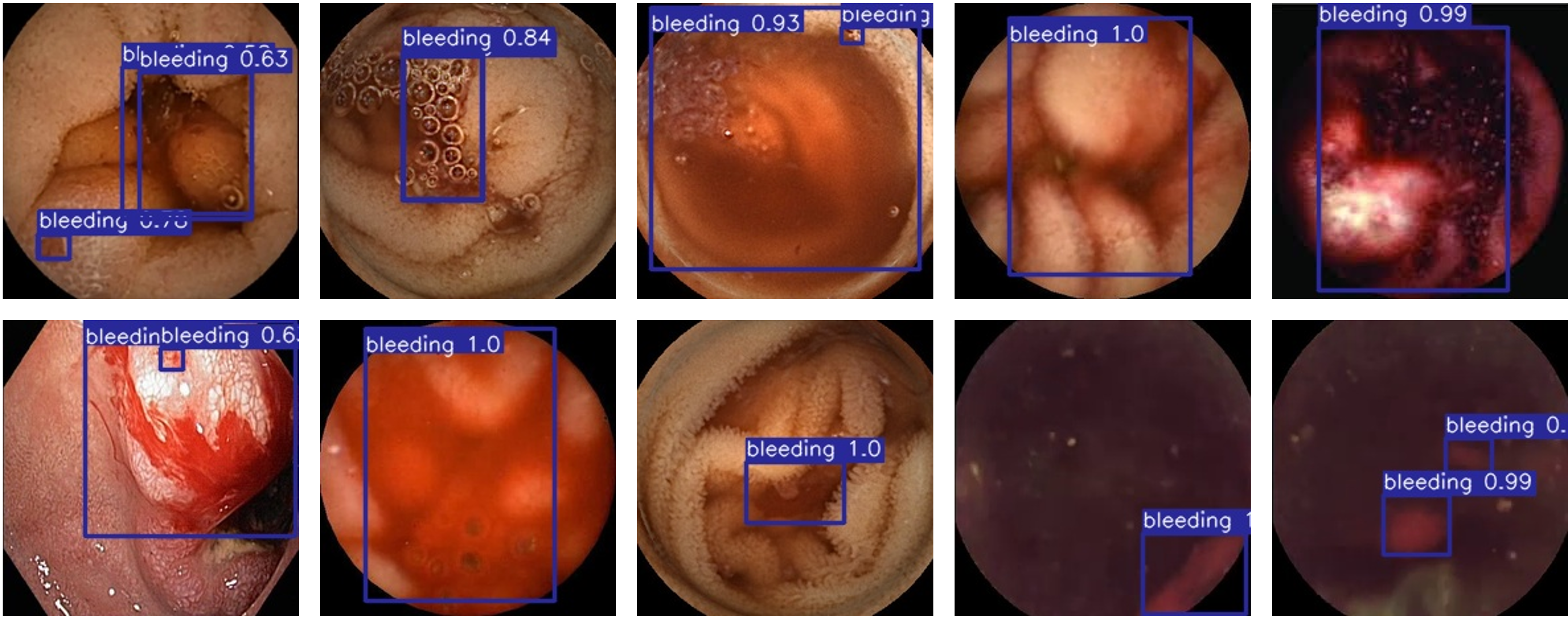}
    \caption{Sample detection images from the validation set.}
    \label{fig:val_set}
\end{figure}

\begin{figure}[htbp]
    \centering
    \includegraphics[width=\linewidth]{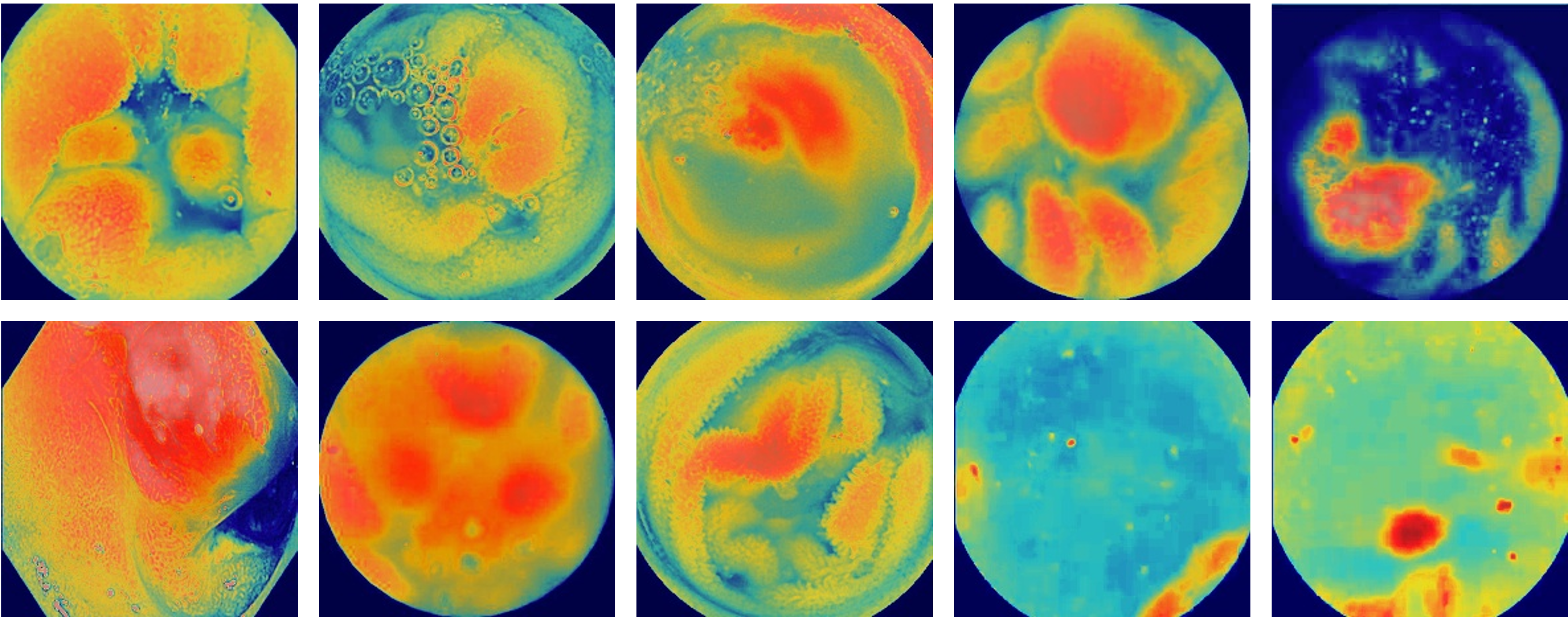}
    \caption{Corresponding Grad-CAM visualization on the Sample validation set of Figure \ref{fig:val_set}.}
    \label{fig:val_set_cam}
\end{figure}

\begin{figure}[htbp]
    \centering
    \includegraphics[width=\linewidth]{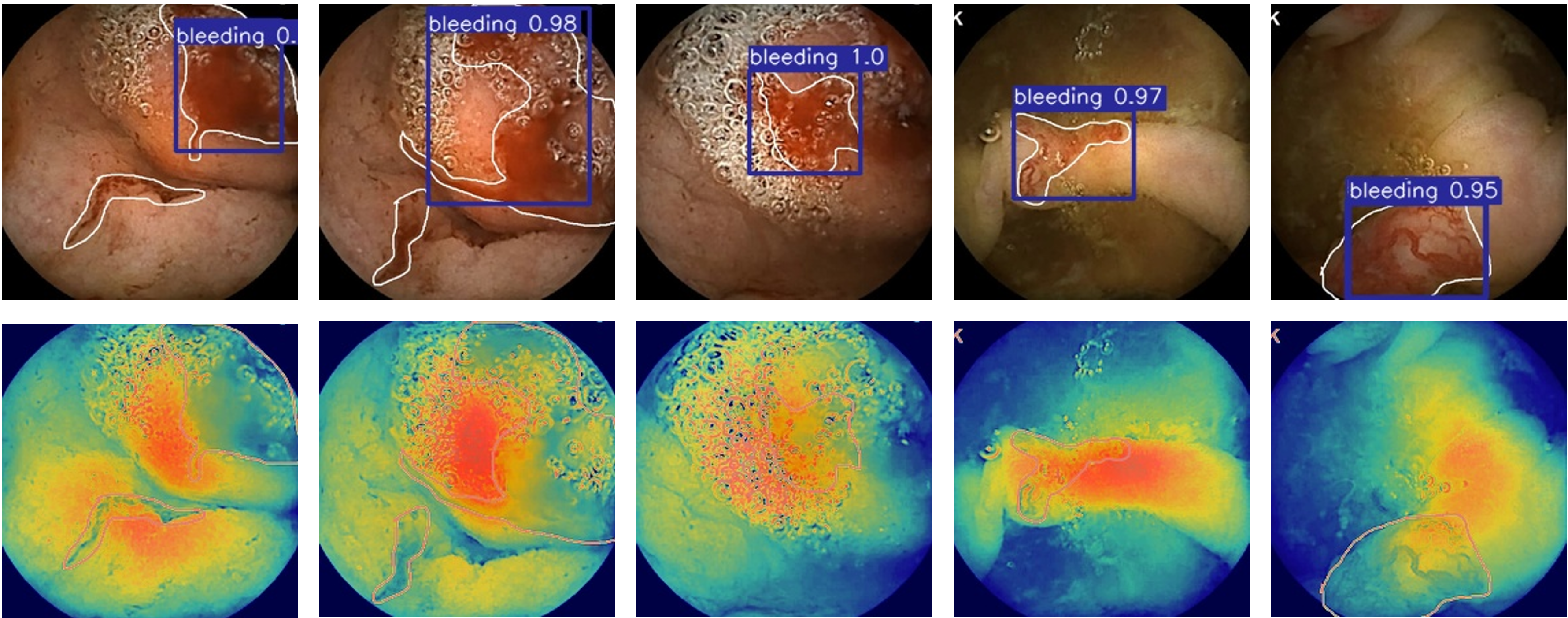}
    \caption{Sample detection images and corresponding Grad-CAM visualization from the test set 1.}
    \label{fig:test_set1}
\end{figure}

\begin{figure}[htbp]
    \centering
    \includegraphics[width=\linewidth]{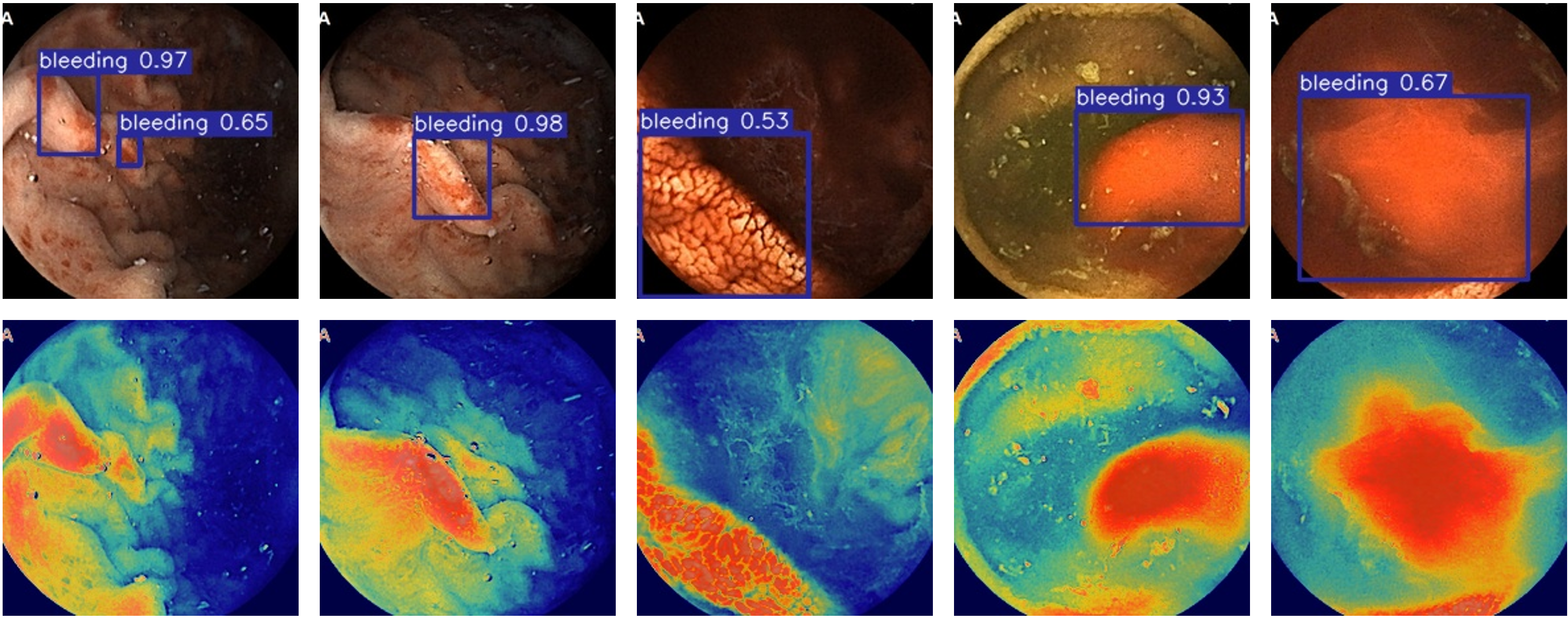}
    \caption{Sample detection images and corresponding Grad-CAM visualization from the test set 2.}
    \label{fig:test_set2}
\end{figure}

\section{Discussion}

Our model effectively utilized the self-attention mechanism for end-to-end object detection and classification with high efficiency. Further, our model yield itself to parallel processing, which is one of the strengths of the transformer-based architectures. However, it requires high computational resources and data. The high data requirement makes training from scratch almost impossible. This limitation, however, can easily be solved using transfer learning, as we have done with our model.

\section{Conclusion}
Our work addresses the challenges of manual analysis in Wireless Capsule Endoscopy (WCE) by developing an end-to-end trainable model for automatic detection and classification of bleeding and non-bleeding frames. Leveraging the DETR architecture our model demonstrates the efficacy in automating the analysis of WCE frames, significantly reducing manual effort while enhancing diagnostic precision. By integrating advanced AI techniques into WCE data interpretation, our work represents a step forward in the adoption of intelligent diagnostic systems, paving the way for more efficient and accurate clinical workflows. 
\label{sec_concl}

\section{Acknowledgments}\label{sec6}
As participants in the Auto-WCEBleedGen Version V1 Challenge, we fully comply with the competition rules as outlined in \cite{hub2024auto} and the challenge website. Our methods have used the training and test data sets provided in the official release in \cite{palakbleedingtrain} and \cite{palakbleedingtest} to report the results of the challenge.

%Please write your list of reference directly in the sample.bib. 
\bibliographystyle{unsrtnat}
\bibliography{main}

\end{document}